# A Convolutional Network Adaptation for Cortical Classification During Mobile Brain Imaging

Benjamin Cichy
*NIWC PAC*
*Intelligent and Sensing Branch*
San Diego, CA
benjamin.cichy@navy.mil

Jamie Lukos
*NIWC PAC*
*Intelligent and Sensing Branch*
San Diego, CA
jamie.lukos@navy.mil

Mohammad Alam
*NIWC PAC*
*Intelligent and Sensing Branch*
San Diego, CA
mohammad.alam@navy.mil

J.Cortney Bradford
*ARL*
U. S. Army Research Laboratory
Aberdeen Proving Ground, MD
jessica.bradford3.civ@mail.mil

Nicholas Wymbs
*NIWC PAC*
*Intelligent and Sensing Branch*
San Diego, CA
nicholas.wymbs@navy.mil

*Abstract*— Deep neural networks (DNN) have become increasingly utilized in brain-computer interface (BCI) technologies with the outset goal of classifying human physiological signals in computer-readable format. While our present understanding of DNN usage for BCI is promising, we have little experience in deciphering neural events from dynamic freely-mobile situations. Using an improved version of EEGNet [1], our goal was to classify cognitive events from electroencephalography (EEG) signals while subjects simultaneously walked on a treadmill, sometimes while carrying a rucksack equivalent to 40% of their body weight. Walking subjects simultaneously performed a visual oddball target detection task, eliciting the P300 event-related potential (ERP), which then served as the DNN classification target. We found the base EEGNet to reach classification levels well above chance, with similar performance to previously reported P300 results. We found performance to be robust to noise, with classification similar for walking and loaded walking, with respect to standard seated condition with minimal movement. With additional architectural search and tuning to the EEGNet model (termed Cog-Neuro, herein; CN-EEGNet), we reached classification accuracy of greater than 95%, similar to previously reported state of the art levels in seated P300 tasks. To our knowledge, these results are the first documented implementation of a DNN for the classification of cognitive neural state during dual-task walking. The classification of one's ongoing cognitive state during a demanding physical task establishes the utility for BCI in complex environments.

## I. Introduction

Electroencephalography (EEG) offers a noninvasive technique for monitoring near-continuous brain activity in humans using an array of electrodes placed on the scalp in a standard configuration. Because EEG offers superior temporal resolution at a relatively low cost, it has become widely used in the development of brain-computer interface (BCI) technology. In practice, BCIs are guided by machine learning models that are capable of developing classification-based rules from EEG input signals. A significant caveat when using EEG is its high susceptibility to noise, whether it be from our own movements and heartbeat to external sources of the surrounding environment. So, while there is a need for BCIs to extend beyond a constrained laboratory setting, it is easy to see that complications may arise when EEG data is collected in more naturalistic settings. We aimed to test the capability of machine learning to classify EEG cognitive state in a mobile environment by measuring P300 detection from subjects as they performed an oddball task while walking on a treadmill.

A particularly useful method for boosting the signal to noise in EEG has been with the application of event-related potential (ERP) paradigms. Using ERPs, cognitive neural states can be inferred through a stimulus-locked EEG waveform. For instance, the canonical P300 waveform represents an elicited response to an infrequently encountered stimulus, which is captured as a positive deflection from baseline approximately 300 milliseconds post-stimulus presentation. The P300 is widely reproducible across subjects and task variations making it a logical choice for some of the first BCI implementations using EEG. The recent development of DNN models for EEG classification of cognitive state has proven most successful when operating on ERPs as input data. Perhaps the best known is the EEGNet model, which has demonstrated strong fine-tuning performance on a variety of ERP classification tasks, including the P300 [2].

Recent advancements in EEG technology have opened the door for the implementation of freely mobile BCIs. Sophisticated noise filtration techniques, such as using independent component analysis (ICA), are able to extract dominant noise spectra and preserve task-relevant signal during tasks that involve dynamic movement, such as walking on a treadmill. For instance, we used ICA noise correction to help quantify changes in EEG cognitive state dynamics in subjects that were walking on a treadmill and also carrying a loaded backpack equal to 40% of their bodyweight [3]. Other work with mobility includes [4], which demonstrated mobile EEG collection on a bicycle, and [5] which included navigating stairs and balancing. With these recent advancements in EEG processing, it is crucial to understand the capability of DNNs for the application of BCIs in freely mobile scenarios.

In recent years convolutional neural networks have (CNNs) have gained wide adoption in EEG signal extraction and classification. Starting with ShallowConvNets [6] and the foundational work of FBCSP [7] that inspired most of today's CNN based on a filter-bank approach for classifying temporal data. The work by Lawhern *et al* with EEGNet [1], reframed FBCSP as a compact convolutional layer system that combined separable 2D convolutional layers to capture the temporal and spatial relationships in EEG signals. More recent works in [8]–[12] have improved on aspects of EEGNet for specific BCI tasking and motor imagery classification.

Our goal was to extend the DNN classification of cognitive EEG events to a freely-mobile scenario using EEGNet and create a new neural network that accounts for mobile user noise. We used high-density array EEG data collected from subjects walking on a treadmill and simultaneously performing an oddball detection dual-task. Oddball tasks provide a reliable means of measuring the P300 ERP in response to the visual detection of the infrequently presented target stimulus. Here, we asked if EEGNet is capable of classifying the P300 ERP during standard dual-task walking, and how classification performance compared to a conventional seated condition with minimal movement. We then asked if classification performance was robust to increased noise through additional physical load (e.g., loaded backpack carriage). Further, we sought to determine if additional tuning and modification of the existing DNN architecture could offer an additional performance boost for the classification of freely mobile EEG cognitive events. Addressing these research questions is a necessary step forward to complement the promising development of mobile EEG technology and the expansive use of DNNs for BCI development.

## II. METHODS

### A. Study Description

The EEG data were collected for an experiment designed to investigate the neural correlates of dual-task performance during loaded walking. Experiment procedures have been previously described in [3]. Briefly, 18 consented subjects (11 men, 7 women; age 26.5±6 yrs.) walked on a treadmill with a rucksack either unloaded or loaded with 40% body weight for 1 hour [13]. 13 subjects were used for this study after post-processing the data in conjunction with subsampling. With data loss during processing, 5 subjects were omitted for this study and further analysis is needed to correct this and reintroduce it in a subsequent dataset. Subjects performed the unloaded and loaded conditions on separate days. At the beginning and end (early and late conditions, respectively) of each walking bout, subjects performed a visual oddball task in which they responded to target stimuli and refrained from responding to non-target stimuli. As a control, subjects were also tested in a seated position prior to and immediately after the walking task (early and late conditions, respectively). We observed no systematic effects of time on task, and pooled early and late conditions accordingly for both walking and seated control. Due to technical challenges involving contaminated signal with artifacts and noise, we are able to report results from 13 of the 18 subjects that performed the experiment.

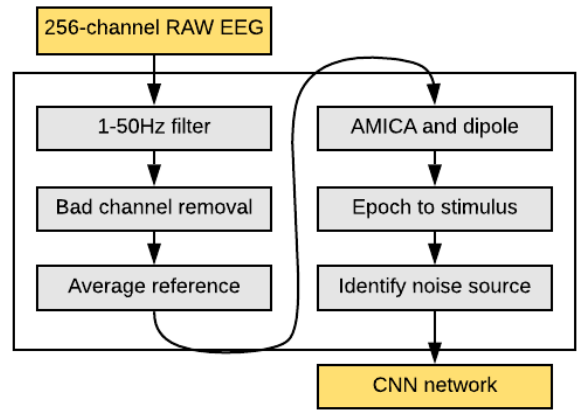

Figure 1 - Data processing on EEG 256-channel input flowchart. The grey boxes shown convey EEGLAB methodology.

### B. Data Balancing

We controlled for overfitting by using a validation subset during model fitting. Using a within-subject model training approach, we partitioned the data by time on task (early/late) for both the movement state (seated/walking) and the weight carriage (loaded/unloaded) conditions. For the Keras (Tensorflow) based model fitting purposes, we split the data for each condition as follows: 70% training, 15% validation, and 15% test. We balanced the test sizes for target and non-target trials by under-sampling the majority class (i.e., non-targets). Additionally, we normalized the data to the early-signal condition length, thereby reducing the overall data-train so it was balanced between early/late seated/unseated conditions. This is equivalent to a 6:1 reduction from the window the signal epoch to the full data sample length.

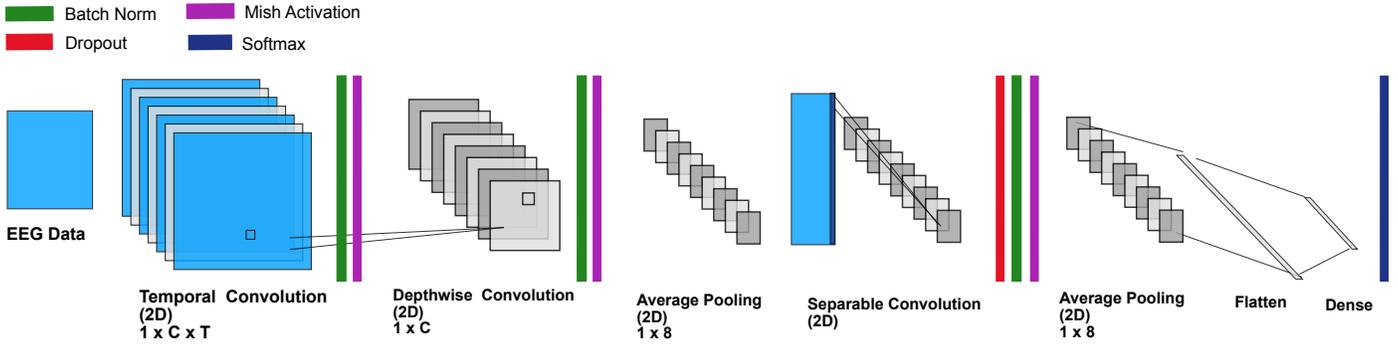

Figure 2 - Layer view of CN-EEGNet Architecture. C = channels. T = time (length) of data.

## C. Architecture

We implemented the EEGNet software general-purpose CNN-based model customized for use on EEG data to handle temporal event-related potentials (ERPs). EEGNet [1] architecture has fewer parameters than a shallow ConvNet [14], while providing flexibility in sampling rate filters (F1, F2). It consists of convolutional layers with non-linear activation function (exponential linear unit), pooling and dropout.

We used EEGNet (v2, 2019) as the base model and then performed modifications to the layers and activation functions as shown in Table 1. We investigated the removal of layers (i.e., dropout) to simplify and accelerate the learning time for the neural network. Dropout [15] is used to prevent overfitting of neural nets to the training set. Further, we implemented three activation functions for tuning hyperparameters: rectified linear unit (ReLU), Swish [16] and Mish [17]. Both Swish and Mish is utilized to improve convergence of network weights during backpropagation, with improved gradient flow through the neural network and. Mish also has the property of being more immune to noisy inputs.

As shown in Table 1 one dropout layer is removed (vs EEGNet) followed by and additional Convolutional 2D layer as shown in layer $\phi_1$. Block $\phi_2$ shows the addition of another Separable 2D convolutional layer This deeper network breaks down additional features without the regularization effect of the dropout layer. The following layers ($\phi_3$) follow the same pattern as EEGNet and output in a softmax layer for the binary classifier. Block $\phi_2$ has a single dropout layer vs two for EEGNet.

Figure 2 shows the resulting data path for the feature extraction system. As described in the above section, the input is a 256-channel set of vectors. The temporal 2D convolution extracts the sequential data points, as well as the Depthwise convolution layer, followed by the Separable convolution for the spatial correlation. Standard average pooling layers compress the representation, and Mish activation is used in between applications of batch normalization as in the blocks $\phi_{1-3}$.

## D. System Tuning

We performed system tuning using both empirical and spot-tuned methods. To accommodate the neural layer style of EEGNet, we used Tensorflow 2.2 (Keras) [18] in legacy mode (1.14). We initially performed early stopping to prevent overfitting. The Keras criteria settings were: *min_delta=0.0001* and *patience=200 (epochs)*. The auto-stop criteria, however, did not work as well as manual tuning from aggregating accuracy values per epoch length. During training, the loss values quickly declined for some subjects {1,2,6}, but not others. The minimum value of the epoch was 100 for these subjects, with 750 needed for the remaining subjects. We took the epoch size of 750 to maximize the overall accuracy at the expense of additional training time. At 750 epochs, there was no evidence of overfitting as the average epoch size was 500-600. Batch size was swept from 32-2048. Literature in [19] pointed to larger batch sizes to improve performance, however, in practice a smaller batch size improved accuracy scores for some subjects {15,21,25,26}. As noted in [20] smaller batches sizes increase per-sample regularization for data.

| # | Layer | Filters/Units | Size | Activation | Output |
|---|---|---|---|---|---|
| $\phi_1$ | Input/Reshape | | (C, T) | | (1, C, T) |
| $\phi_1$ | Conv 2D | $F_1$ | (1,C,T) | Linear | ($F_1$, C, T) |
| $\phi_1$ | Batch Normalization | | - | | ($F_1$, C, T) |
| $\phi_1$ | Activation | | - | Mish | |
| $\phi_1$ | DepthwiseConv2D | $D * F_1$ | (C,1) | | ($D*F_1$, 1, T) |
| $\phi_1$ | Batch Normalization | | - | | ($D*F_1$, 1, T) |
| $\phi_1$ | Activation | | - | Mish | ($D*F_1$, 1 T) |
| $\phi_1$ | Average Pooling 2D | | (1,8) | | ($D*F_1$, 1, T/4) |
| $\phi_2$ | SeparableConv2D | $F_2$ | (1,16) | Linear | ($F_2$, C, T/4) |
| $\phi_2$ | Spatial Dropout | | - | | ($F_2$, C, T/4) |
| $\phi_2$ | SeparableConv2D | $F_2$ | (1,16) | | ($F_2$ C, T/4) |
| $\phi_2$ | Batch Normalization | | - | | ($F_2$, C, T/4) |
| $\phi_2$ | Activation | - | - | Mish | ($F_2$, C, T/4) |
| $\phi_2$ | Avg Pooling 2D | | (1,8) | | ($F_2$, 1, T/32) |
| $\phi_3$ | Flatten | | | | ($F_2*(T/32)$) |
| $\phi_3$ | Dense | N | | | N |
| $\phi_3$ | Activation | | | Softmax | N |

Table 1- CN-EEGNet architecture, where C = number of channels, T = number of time points, F1 =number of temporal filters, D = depth multiplier (number of spatial filters), F2 = number of pointwise filters, and N = number of classes, respectively.

*E. Hyperparameters and tuning*

We used HyperOpt [7] to optimize model parameters in order to tune libraries for a basic sweep. We selected the following parameters for tuning: norm-rate, spatial filter number (D), number of temporal filters ($F_1$), number of pointwise filters ($F_2$), dropout rate, regularization rate, and optimizer type (*adam, sgd, adagrad, adadelta, adabelief [21], adamax, nadam*). We swept with the default values in Table 2 until we narrowed our search to three potential optimizers (*adabelief, adam, and nadam*). These optimizers had stable accuracy values across all subjects and did not overfit.

The three optimizers were further tuned across their respective learning rate parameter, beta, and epsilon and learning rate (*lr=0.0009, beta_1=0.9, beta_2=0.999, epsilon=1e-7*). Here, the design parameters correspond to: $F_1$ = number of temporal filters, $F_2$ = pointwise filters, $D$ = number of spatial filters as shown in Table 1.

| Value Type | F1 | F2 | D | Dropout Rate | Kernel Length | Norm Rate | Opt |
|---|---|---|---|---|---|---|---|
| Default | 8 | 16 | 2 | 0.25 | 64 | 0.25 | adam |
| Optimized | 32 | 16 | 8 | 0.25 | 128 | 0.25 | adam |

Table 2- Baseline EEGNet parameter settings, HyperOpt and empirically tuned base settings as generalized for the subsequent networks.

| Value Type | F1 | F2 | D | Dropout Rate | Kernel Length | Norm Rate | Opt |
|---|---|---|---|---|---|---|---|
| Default | 8 | 16 | 2 | 0.25 | 64 | 0.25 | adam |
| Optimized | 16 | 16 | 2 | 0.15 | 64 | 0.17 | adam |

Table 3- Optimized values for CN-EEGNet for the walking dataset. The recommended values did not vary significantly with tuning.

### III. RESULTS

*A. EEGNet P300 Classification: Dual-Task Walking*

We observed that EEGNet performed well on the classification of P300 ERPs from subjects while they performed an oddball detection while simultaneously walking on a treadmill at a comfortable pace. Depicted in Figure 3A, EEGNet reached a mean accuracy of 89.04% ($SD$ = 12.6) for the unloaded walking condition, which was better than or at least similar to with seated unloaded performance of the oddball task ($M$ = 86.29; $SD$ = 14.5). This result shows that EEGNet is more than capable of performing within-subject classification on data acquired by means of mobile EEG platforms, such as walking on a treadmill.

*B. EEGNet P300 Classification: Walking Under Heavy Load*

We performed a manipulation in which subjects on half of the trials, performed dual-task walking while also carrying a loaded backpack equal to 40% of their body weight. Given that EEG signal is susceptible to sources of noise including large amplitude signals from the muscles, we tested if EEGNet is robust to additional muscle activations in response to the additional weight of the backpack. Depicted in Figure 3A, we found that EEGNet performed well on the loaded walking condition ($M$ = 91%; $SD$ = 11.76), with performance on par or better than the seated loaded control condition ($M$ = 86.5%; $SD$ = 9.96). Interestingly, we observed better accuracy for the loaded walking condition with respect to the unloaded walking condition (91% vs 89.04%). These results suggest that when combined with suitable preprocessing, such as ICA noise extraction (e.g., EEGLab's AMICA), CN-EEGNet is a strong candidate for classification of ERPs acquired under a freely mobile EEG configuration.

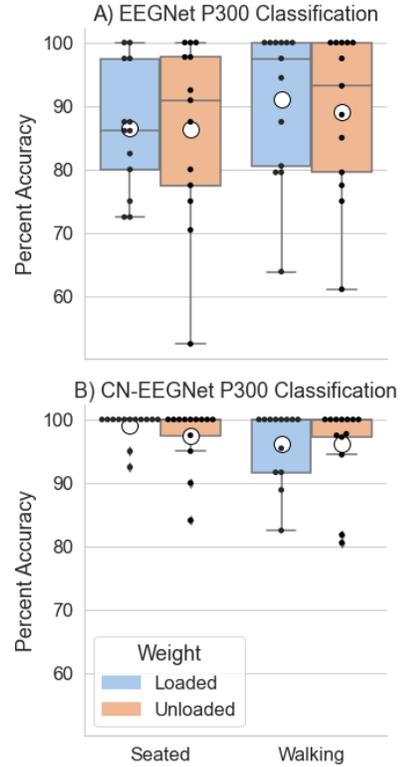

Figure 3 - EEGNet (A) vs CN-EEgNet (B) across seated and walking (loaded/unloaded) modalities.

*C. CN-EEGNet vs EEGNet P300 Classification*

We tested whether additional architectural search as well as tuning could provide an appreciable difference in classification performance of EEGNet. In our modified EEGNet, termed CN-EEGNet, the removal of the pooling layers allowed for greater exposure of the EEG signal at the input layer. We applied several activation functions, collectively Swish [16], Mish and Relu. These activation functions were substituted for the standard 'Elu' function in EEGNet. Indeed, we found that CN-EEGNet greatly outperformed EEGNet on all phases of classification (Figure 3B). Averaging across all conditions, classification accuracy for CN-EEGNet was 97.18%, whereas EEGNet was 88.21%, with a notable difference also found in the standard deviation of accuracy (5% vs 12.2%, CN-EEGNet and EEGNet, respectively). Taking a closer look, we found that CN-EEGNet classification performed equally well for walking without a load ($M$ = 96.09%; $SD$ = 6.84%) and walking with a heavy load ($M$ = 96.17%; $SD$ = 5.76%). These accuracy levels were comparable to CN-EEGNet performance during the seated control conditions (unloaded: M = 99.43%; $SD$ = 4.99%; loaded: $M$ = 99.04%; $SD$ = 2.4%). The CN-EEGNet extension to EEGNet demonstrates measurable improvement in classification performance above the current state of the art. These improvements to the general framework of EEGNet suggest that DNNs like our CN-EEGNet are capable of robust within-subject classification performance of freely mobile EEG data.

*D. P300 Classification: Extended Model Comparison*

We selected additional commonly utilized CNNs for additional comparison to understand if classification of freely mobile ERP epochs was specific to EEGNet and similarly derived architecture (CN-EEGNet). Illustrated in Figure 4, models with the exception of ConvNet-Deep (M: 65.06%; SD: 7.1%) maintained accuracy near state of the art across subjects (*ConvNet-Shallow*, *M*: 94.05%, *SD*: 4.33%; *EEGNet*, 88.21%, *SD*: 6.19%; *CN-EEGNet:* 97.18%, *SD*: 2.45%). ConvNet-Shallow and CN-EEGNet were highest in similarity, suggesting that relatively shallower networks are perhaps better suited for classification of EEG signals with relatively well-defined components of interest – in this case the P300. Deeper network architectures such a Resnet[22] use skip connections for gradient flow – whereas CN-EEGNet is balanced in depth, similar to EEGNet. Lastly, the variability of within-subject classification in Figure 4 highlights the growing need to understand what intrinsic features of the EEG signal are most salient for classification, and ultimately what model is best at capturing these potential features in a between-subject basis (in our case, CN-EEGNet).

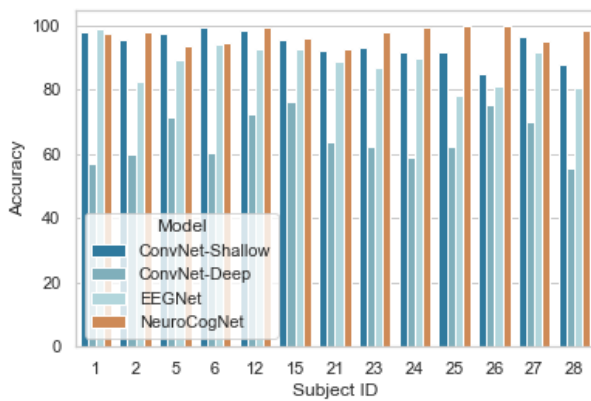

Figure 4. - Comparison of within-subject P300 classification for several commonly used CNNs (including EEGNet and CN-EEGNet).

## IV. SUMMARY

In this paper, we have investigated the impact of using a CNN for detecting P300 ERP waveforms for BCI applications in a mobile platform. We developed a modified version of EEGNet (CN-EEGNet), which improved upon previous work, by not only increasing classification accuracy, but also reducing variability of within-subject classification of the P300 during high noise conditions (i.e., walking with and without a heavy load). Were able to both increase accuracy, as well as reduce variability.

As seen in with experiments by other EGG classification DNNs against MOAB[14], the evoked response by certain subjects can differ in ways that affect general classification. Our network, without much optimization (HyperOpt or similar hyperparameter tuning software), can capture and classify ERPs regardless of the additional artifacts induced by a freely-mobile setting. This work represents a positive step towards increasing DNN accuracy, and with relatively low between-subject variability, suggests that intra-subject accuracy of EEGs is viable approach.

ACKNOWLEDGMENT

The authors would like to thank Sandia Rawal and Joseph McArdle for their help with data collection and the members of the ARL Human Research and Engineering Directorate for their feedback on data analysis.